# An Introduction to a New Text Classification and Visualization for Natural Language Processing Using Topological Data Analysis

Naiereh Elyasi

Department of Mathematics, Faculty of Mathematics and Computer Science, Kharazmi University, Tehran, Iran, Email: elyasi82@khu.ac.ir

Mehdi Hosseini Moghadam

Department of Mathematics, Faculty of Mathematics and Computer Science, Kharazmi University, Tehran, Iran, Email:m.h.moghadam1996@gmail.com

**Abstract:** Topological Data Analysis (TDA) is a novel new and fast growing field of data science providing a set of new topological and geometric tools to derive relevant features out of complex high-dimensional data. In this paper we apply two of best methods in topological data analysis, "Persistent Homology" and "Mapper", in order to classify persian poems which has been composed by two of the best Iranian poets namely "Ferdowsi" and "Hafez". This article has two main parts, in the first part we explain the mathematics behind these two methods which is easy to understand for general audience and in the second part we describe our models and the results of applying TDA tools to NLP.

## 1 Introduction

### 1.1 Authorship Attribution

These days internet is full of text based information including (news, messages, comments, ...), in order to analyze this huge amount of unstructured data we need to classify them. The process of assigning labels to some text based on their content is called text classification. Authorship attribution is one of the main branches in text classification which tries to identify the author of a given text based on its content.

We can divide authorship attribution into two main methods:
– Non-Semantic: This method tries to identify the author based on the length of words and sentences and the vocabulary used in a given text[10][16].
– Semantic: This method considers the structure of the language based on its semantic analysis[4][5][26].
In this article we try to do authorship attribution based on a new state-of the-art method called topological data analysis.

### 1.2 Topological Data Analysis

Data is every where, if you just think about our daily life you could find out how much information and data is produced every day by a single person. With ever growing amount of data and advances in technology we face with the world of information overload. In order to make our life easier and having a good sense of what is happening in our world, what the threats and opportunities are, we need to collect and analyze this huge amount of data. Data science is the field of science in which it is grappling with huge and messy amount of data in order to turn it to something useful. One of the main difficulties in data science is to deal with high dimensional data and transform it into data with less dimensionality in order to make it easier for analyzing. Topological data analysis (TDA) is one of the newest and fast growing branches of data science which is trying to analyze data by studying its shape and also reducing the dimensionality of data. TDA is based on two very important branches "Statistics" and "Algebraic Topology", due to its methodology TDA can solve some serious problems in data science, because it is a clustering method that is robust to noise. The goal of TDA is to reduce the dimensionality of high dimensional data and also analyzing the topological structure or shape of data and finally clustering complex data. Since the nature of data is random, TDA tools has been extended by the statistical and combinatorial concepts. TDA also provides innovative data mining methods that can improve the efficiency of machine learning techniques. TDA has two main methods "Persistent Homology" and "Mapper". In Persistent homology, a filtration of combinatorial objects, simplicial complexes, is constructed and then main topological structures of data is derived. Some visualization tools such as "Persistent Diagram","Barcode" and " Persistent Landscape" are invented to indicate the main topological features of data. Persistent homology has been previously used in pulse pressure wave analysis[7], analyzing 3D images[9], brain networks[12], image analysis[6], Path Planning[1]. The idea behind Mapper is to reduce a highdimensional and noisy data set into a combinatorial object (simplicial complex). Such an object tries to encapsulate the original shape of the high-dimensional data. TDA Mapper has been previously applied to data in breast-cancer patients[19], Text Representation for Natural Language Processing[29], Text mining[29][8][20], topic detection in twitter[27] and clinical data[19][23][21][18][11]. In this paper first we introduce some mathematical backgrounds about "Group Theory", "Simplicial Complex" and "Homology". Next we introduce "Persistent Homology" algorithm, then we apply this method to authorship attribution (dataset of poems) and analyze the results. Afterwards we introduce a novel method in TDA called "Mapper". We apply mapper to authorship attribution (dataset of poems) and analyze the result as a simplicial complex which is interactive and can be quantified in several ways using statistics. After wards we introduce "Persistent Homology" algorithm, next we apply this method to authorship attribution (dataset of poems) and analyze the results. In last section we compare the results of applying persistent homology and mapper on dataset of poems.

## 2  Preliminaries

### 2.1 Group Theory

**Definition 1** *A group is a pair* $(G, *)$ *which* $*$ *is a binary operation on set a G and an identity element* $\underline{e}$ *that satisfies:*
• $(a * b) * c = a * (b * c)$ *for all a,b,c,* $\in G$ *(associativity)*
• $e * a = a$ *for all* $a \in G$.
• *for every* $a \in G$, *there is an element* $b \in G$ *such that* $b * a = e$
*(b in the third part of definition is called the inverse of* a *and is denoted by* $a^{-1}$).

**Example 2** *The integers* Z *under addition* $+$ *is a group.* From now on we use notation $G$ instead of $(G, *)$.

**Definition 3** *A subset* $H \subseteq G$ *of a group G with operation* $*$ *is a subgroup of G if* $(H, *)$ *is itself a group.*

**Definition 4** *A group G is abelian if* $a * b = b * a$ *for all elements a,b* $\in G$.

**Definition 5** *Two groups G and H are said to be isomorphic if there is bijective function* $\theta : H \rightarrow G$ *such that:*

$$\theta (g_1 g_2) = \theta (g_1) \, \theta (g_2) \tag{1}$$

*for all* $g_1, g_2 \in G$. *The function* $\theta$ *is called an isomorphism. If* $\theta$ *satisfies only the equation* 1, *then* $\theta$ *is named a homomorphism.*

**Definition 6** *If* $\theta : H \rightarrow G$ *is a group homomorphism, the kernel of* $\theta$ *is defined by:*

$$\mathrm{Ker}(\theta) = \{a \in G \mid \theta(a) = e\}$$

*and the image of* $\theta$ *is defined by:*

$$\mathrm{Im}(\theta) = \{ \, \theta(a) \mid a \in G \}$$

**Definition 7** *Let H be a subgroup of a group G. If* $g \in G$, *the right coset of H generated by g is*

$$\text{Hg} = \{\text{hg} \mid \text{h} \in \text{H}\}$$

*Similarly the left coset of H generated by g can be defined*

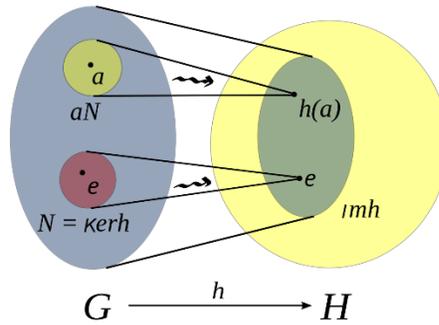

**Fig. 1:** Representation of Ker and Image function h

**Definition 8** *The cosets* {aH | a ∈ G} *under the operation* ∗ *form a group, called the quotient group* H/G.

**Definition 9** *Let S ⊆ G, the subgroup generated by S denoted by* ⟨S⟩ *is the subgroup of all elements of G that can expressed as the finite operation of elements in S and their inverses.*

**Definition 10** *The rank of a group G is the size of the smallest subset that generates G.*

## 2.2 simplicial Complex and Homology

**Definition 11** *Let $u_0, u_1, ..., u_k$ be points in $R^n$. A point $x = \sum_{i=1}^{k} \lambda_i u_i$ with each $\lambda_i \in R$, is an affine combination of the $u_i$ if $\sum_{i=1}^{k} \lambda_i = 1$.*

**Definition 12** *The k+1 points are affinely independent if the k vectors $u_i - u_0$, for $1 \leq i \leq k$, are linearly independent.*

**Definition** 13 *An affine combination $\sum_{i=1}^{k} \lambda_i$ is called a convex combination if each $\lambda_i$ are non-negative.*

**Definition 14** *The convex hull of $U = \{u_0, u_1, ..., u_k\}$ is the set of all convex combonations of U .*

**Definition 15** *A k-simplex σ is the convex hull of k+1 affinely independent points $u_0, u_1, ..., u_k$. So we can see in figure 2 that* 0-*simplex is a vertex,* 1-*simplex is an edge,* 2-*simplex is a triangle, and* 3-*simplex is a tertrahedon.*

Now we want to define a special union of some simplexes that is called simplicial complex. The intuitive definition of simplicial complex *K* is that if a simplex is in *K*, all of its faces need to be in *K*, too. In addition, the simplexes have to be glued together along whole faces or be separate. The figure 3 showes some useful examples.

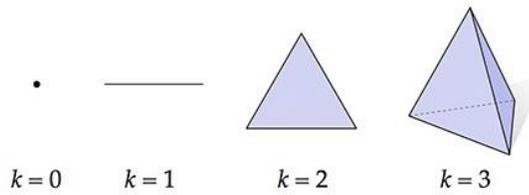

**Fig. 2:** The figure shows $k$-simplexes for $k = 0;1;2;3$

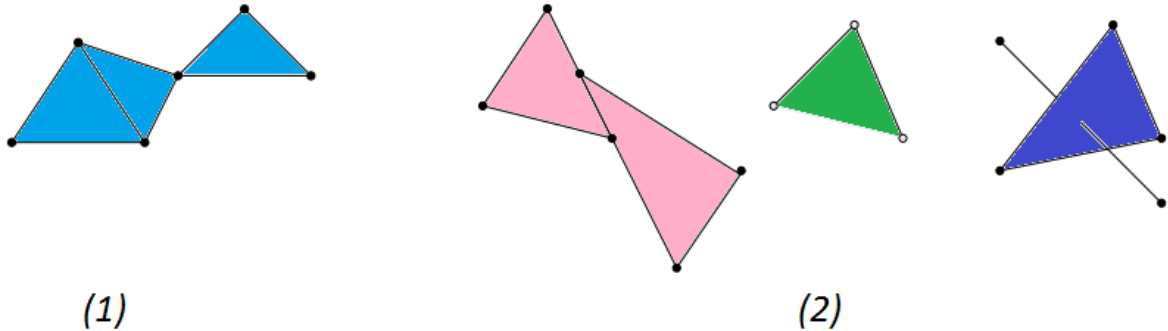

*(1)*                                        *(2)*

**Fig. 3:** figure number 1 showes a simplicial complex but figures in number 2 are not simplicial complexes

**Definition 16** *A face of simplex $\sigma$ induced by { $u_i$} is the convex hull of a nonempty subset of the { $u_k$}, and it is called proper face if the subset is not the entire set. We some times write $\tau \leq \sigma$ if the $\tau$ is a face of s and $\tau < \sigma$ if it is a proper face of $\sigma$ .*

**Definition 17** *A simplicial complex K is a finite collection of simplexes such that first $\sigma \in K$ and $\tau \leq \sigma$ implies $\tau \in K$ , and second $\sigma, \sigma_0 \in K$ implies $\sigma \cap \sigma_0$ is either empty or a face of both.*

**Definition 18** *The p-th chain group of a simplicial complex K is $(C_p(K), +)$, the free Abelian group on the oriented p-simplices, where $[\sigma] = - [t]$ if $\tau = \sigma$ furthermore $\sigma$ and $\tau$ have different orientations. An element of $C_p(K)$ is called a p-chain, denoted by $\sum_{i=1}^{k} \lambda_i \, [\sigma_i]$  where $\lambda_i \in \mathbb{Z}$, $\sigma_i \in K$.*

**Definition 19** *Let K be a simplicial complex and $\sigma \in K$, $\sigma = [v_0, v_1, ..., v_k]$. The boundary homomorphism $\partial_k \colon C_k(K) \rightarrow C_{k-1}(K)$ is defined as follows:*

$$\partial_k \sigma = \sum_i (-1)^i \, [v_0, v_1, ..., \widehat{v_i}, ..., v_n]$$

*Where $\widehat{v_i}$ indicates that $v_i$ is deleted from the sequence. It is easy to check that $\partial_k$ is well defined, that is, $\partial_k$ is the same for every ordering in the same orientation.*

**Example 20** *The boundary of the oriented triangle in the figure 4 is as follows:*
$$\partial[a,b,c] = [b,c] - [a,c] + [a,b] = [b,c] + [c,a] + [a,b].$$

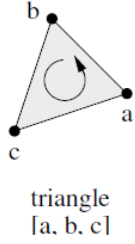

triangle
[a, b, c]

**Fig. 4:** A 3-dimensional simplicial complex

**Definition 21** *The k-th cycle group is $Z_k = Ker\partial_k$. A chain that is an element of $Z_k$ is named a k-cycle. The k-th boundary group is $B_k = Im\partial_{k+1}$. A chain that is an element of $B_k$ is called a k-boundary.*

**Lemma 1** *Fundamental lemma of homology: For every (p+1)-chain d we have $\partial_k\ \partial_{k+1}d = 0$.*

From the last lemma, we know that the group of boundaries form a subgroups of the cycles group and then we can take quotients. In other words, we can partition each cycle group into classes of each cycles that differ from each other by boundaries. This leads to the notion of homology groups and their ranks, which we now define and discuss.

**Definition 22** *The p-th homology group is the p-th cycle group modulo the p-th boundary group, denoted by $H_p = Z_p/B_p$. The p-th betti number, denoted by, ß$_p$ is the rank of group $H_p$.*

*A Vietoris-Rips complex of diameter ε is the simplicial complex that is defined as follows:*

$$VR(\varepsilon) = \{\ \sigma\ /\ diam(\sigma) \le \varepsilon\ \}$$

*where $diam(\sigma)$ is the largest possible distance of points in $\sigma$ .*

**Definition 24** *A sequence of nested simplicial subcomplexes*

$$\emptyset \subseteq K_0 \subseteq K_1 \subseteq K_2 \subseteq \cdots$$

*is called a filtration.*

**Definition 25** *Consider the following filteration*

$$\emptyset \subseteq K_0 \subseteq K_1 \subseteq K_2 \subseteq \cdots$$

*. Naturally we will have a sequence of homomorphisms induced by inclusions*
$$H_p(K_0) \to H_p(K_1) \to H_p(K_2) \to \cdots$$

*We say that a p-th homology class $[c]$ is born at $K_i$, if $[c] \in K_i$ but $[c] \notin K_{i-1}$ and we say $[c]$ dies in $K_i$, if it merges with a class that is born earlier i.e $[c] \in K_{i-1}$ and $[c] \notin K_i$*

## 3 Persistent Homology

Persistent homology tries to find and track homological groups and holes with the help of filtrations. In this section we try to explain the concept and algorithm of persistent homology and then we apply it to text classification. A filtered simplicial complex with its boundary functions is called a persistent complex. For an increasing sequence of positive real numbers we attain a persistent complex. Our goal is to analyze topological invariants of a point cloud data set by analyzing its persistent complex. some normal representation for persistent homology are barcode, persistent diagram and lanscape . A barcode represents each persistent generator(classes that generate *p*-th homology group) with a horizontal line beginning at the first filtration level where it appears, and ending at the filtration level where it disappears, while a persistence diagram plots a point for each generator with its x-coordinate the birth time and its y-coordinate the death time. A visual representation of persistent homology is persistent diagram.

**Definition 26** *The p-persistent diagram D of a filtration*

$$\emptyset \subseteq K_0 \subseteq K_1 \subseteq K_2 \subseteq \cdots$$

*is defined as follows.*
*Let* $\mu_p^{ij}$ *be the number of independent p- dimensional classes that are Born in $K_i$ and die entering $K_j$, then D is obtained by drawing a set of points $(i, j)$ with multiplicity* $\mu_p^{ij}$ *, where the diagonal is added with infinite multiplicity.*

For comparing two persistent diagrams some metrics are defined that two of most important of them are bottleneck and wasserstein distances.

**Definition 27** *Let $D_1, D_2$ be two persistent diagrams and B be the set of all bijective functions $\varphi : D_1 \rightarrow D_2$. If $\| \cdot \|_\infty$ be the supremum norm, then the bottleneck distance between two persistent diagrams $D_1, D_2$ denoted by $W_\infty (D_1, D_2)$ is defined as follows.*

$$W_\infty (D_1, D_2) = \inf sup_{\varphi \in B_{x \in D_1}} \| x - \varphi(x) \|_\infty$$

**Definition 28** *Let $D_1, D_2$ be two persistent diagrams and B be the set of all bijective functions $\varphi : D_1 \rightarrow D_2$, then the Wasserstein distance between two persistent diagrams $D_1, D_2$ denoted by $W_p (D_1, D_2)$ is defined as follows.*

$$W_p (D_1, D_2) = [inf_{\varphi \in B} \sum_{x \in D_1} \| x - \varphi(x) \|_\infty^p]^{\frac{1}{p}}$$

Since it is very hard to analyze the information about homological groups and holes we can use a visualization method called "Barcode", the idea is as follows:
if a hole appears in $\varepsilon_{t1}$ we start to draw a line where the begining of the line is at $\varepsilon_{t1}$ in x-axis and if die in $\varepsilon_{t2}$ , we stop drwing the line and end of the line would be at $\varepsilon_{t2}$ . Persistent landscape is another method introduced by bebunik[2] to visualize persistent homology .

**Definition 29** *The persistence landscape is a function $\lambda$: N×R $\rightarrow \overline{R}$ where $\overline{R}$ denotes the extended real numbers $[-\infty, \infty]$. Alternatively, it may be thought of a sequence of functions $\lambda_k$ : R $\rightarrow$ R , where $\lambda_k (t) = \lambda (k, t)$. Define $\lambda_k (t) = \sup\{m \geq 0 \mid \beta^{(t-m), (t+m)} \geq k\}$ where $\beta^{i,j}$ is the dimension of group $H_i / H_j$.*

The graph of landscape indicates persistent and non-persistent betti numbers, for example the support of persistent landscape denotes non-persistent betti numbers and the maximum of landscape graph indicate the most persistent betti number.

### 3.1 Persistant Homology Algorithm
Let P be a point cloud data. First we construct the Vietoris-Rips complex for P as follows:
consider an increasing sequence of positive real numbers $\varepsilon_1 \leq \varepsilon_2 \leq \varepsilon_3 \leq \cdots$, then we construct a cover of circles with centers of points in P and diameter $\varepsilon_1$, so we have as many circles as the number of data points in point cloud data, next we draw an edge between the center of each two circle which have any intersection and therefore we have a simplicial complex VR($\varepsilon_1$). We do the same process for all $i = 1,2,3,\ldots$ as a result we have a filtration of complexes VR($\varepsilon_i$). Since it is very hard to analyze the information about homological groups and holes we can use a visualization method called "Barcode", the idea is as follows:

if a hole appears in $\varepsilon_{t1}$ we start to draw a line where the begining of the line is at $\varepsilon_{t1}$ in x-axis and if die in $\varepsilon_{t2}$, we stop drwing the line and end of the line would be at $\varepsilon_{t2}$.

## 3.1 Persistant Homology Algorithm

Let P be a point cloud data. First we construct the Vietoris-Rips complex for P as follows:

consider an increasing sequence of positive real numbers $\varepsilon_1 \leq \varepsilon_2 \leq \varepsilon_3 \leq \cdots$, then we construct a cover of circles with centers of points in P and diameter $\varepsilon_1$, so we have as many circles as the number of data points in point cloud data, next we draw an edge between the center of each two circle which have any intersection and therefore we have a simplicial complex VR($\varepsilon_1$). We do the same process for all $i = 1,2,3,\ldots$ as a result we have a filteration of complexes VR($\varepsilon_i$). Since it is very hard to analyze the information about homological groups and holes we can use a visualization method called "Barcode", the idea is as follows:

if a hole appears in $\varepsilon_{t1}$ we start to draw a line where the begining of the line is a $\varepsilon_{t1}$ in x-axis and if die in $\varepsilon_{t2}$, we stop drwing the line and end of the line would be at $\varepsilon_{t2}$.

## 3.2 Results

Persistent homology has already been implemented in R packages like "TDA" and "TDAstatus", in this article we used these two packages for text classification. In this part we used the textual data (poems) of two iranian poets (Hafez and Ferdowsi), the data set gathered from "Shahnameh"[1] and "Ghazaliat-e-Hafez". It includes about 8000 hemistich from each book. After preprocessing we fed the data to tf-idf algorithm in order to make document term matrix, next we fed the document term matrix to persistent homology algorithm. First we sketch presistent diagram, barcode and persistent landscapes for a sample of Ferdowsi poems including 1000 hemistich that is shown in figure 5. We also devided 8000 hemistich of "hafez" to 8 parts with lenght 1000, then we computed persistent diagram and first landscape of each part. Finally we sketched the mean landscape of these parts. We did same work for 8000 hemistich of "Ferdowsi". At last step we computedWasserstein distances between persistent Diagrams of correspondance parts of "hafez" and "Ferdowsi" poems. We bring these results as follows.

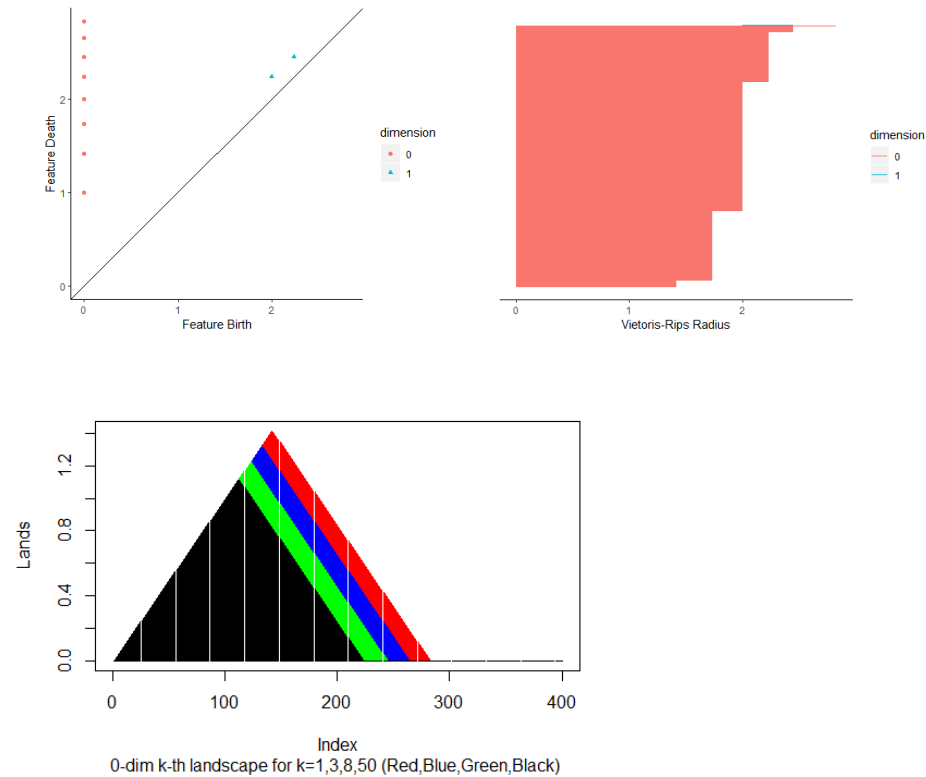

0-dim k-th landscape for k=1,3,8,50 (Red,Blue,Green,Black)

1 An epic book from Ferdowsi

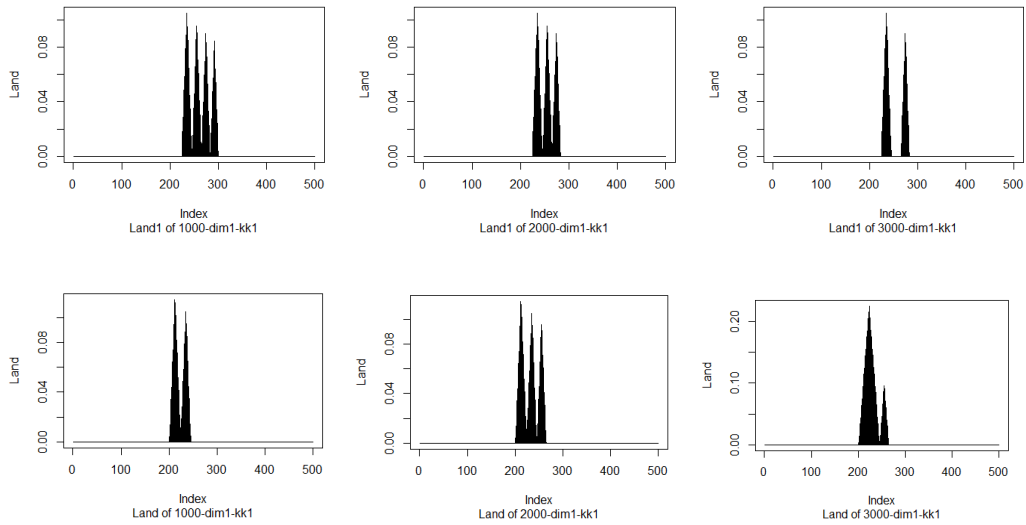

**Fig. 6:** First row shows persistent of land scapes of first there parts of Hafez poems and second row shows persistent of land scapes of first there parts of Ferdowsi poems

We bring the results of computing Wasserstein dictances between different partsof poems in tables 1 and 2 as follows.

| Different parts of poems of Ferdowsi and Hafez | Distance of Dim 0 | Distance of Dim 1 |
|---|---|---|
| part1 of Ferdowsi and part1 of Hafez | 449.4274 | 0.6881788 |
| part2 of Ferdowsi and part2 of Hafez | 82.16098 | 0.2928399 |
| part3 of Ferdowsi and part3 of Hafez | 82.84308 | 3.765808 |
| part4 of Ferdowsi and part4 of Hafez | 553.3845 | 2.078418 |
| part5 of Ferdowsi and part5 of Hafez | 118.2316 | 3.312112 |
| part6 of Ferdowsi and part6 of Hafez | 141.321 | 0.5248254 |
| part7 of Ferdowsi and part7 of Hafez | 74.28296 | 3.282468 |
| part8 of Ferdowsi and part8 of Hafez | 387.2628 | 0.4503871 |

**Table 1:** Wasserstein0-distances and 1-distances is computed for coresponding parts of Hafez and Ferdowsi poems

| Different parts of Ferdowsi poems | Distance of Dim 0 | Distance of Dim 1 |
|---|---|---|
| part1 and part2 | 146.1085 | 0.3634205 |
| part2 and part3 | 5.64018 | 3.405932 |
| part3 and part4 | 310.8692 | 1.946319 |
| part4 and part5 | 278.3973 | 1.427155 |
| part5 and part6 | 38.93984 | 3.003388 |
| part6 and part7 | 14.5807 | 3.266145 |
| part7 and part8 | 16.77798 | 3.322824 |

**Table 2:** Wasserstein distances between consequences parts of Ferdowsi poems

# 4    Mapper Algorithm

The Mapper algorithm was introduced by Singh, Mmoli and Carlsson[25] as a geometrical tool to analyze and visualize datasets.

Here we use below notation to explain the mathematics of mapper method.

| | Symbol | Explanation |
|---|---|---|
| **1** | $X$ | Underlying space of point cloud data ($R^n$ for some $n \in N$) |
| **2** | $P$ | Point cloud data ( $P \subseteq X$) |
| **3** | $Y$ | Parameter space (usually $Y$ is equal to real numbers) |
| **4** | $F$ | Filter function ( $F: \rho \subseteq X \longrightarrow Y$ ) |
| **5** | $\Gamma$ | Range of $F$ restricted to $\rho$ |
| **6** | $U$ | A covering of $\rho$ ($\rho \subseteq \bigcup_{\alpha \in U} \alpha$) |
| **7** | $\xi$ | Collection of subintervals of $\Gamma$ which overlap ($\xi$ covers $\Gamma$ ) |

*__Table__ 3: Notation*

• First we start with a suitable filter function $F: \rho \subseteq X \longrightarrow Y$

• Then we find the range of $F$ restricted to $\rho$ and call it $\Gamma$ . Then partition $\Gamma$ into subintervals $\xi$ in order to create a covering of $\rho$ with $F^{-1}$ in the next step

• For every subinterval $\xi_i \in \xi$ we make the following set $X_i = \{ x \mid F(x) \in \xi_i \}$, the set $U = \{X_i\}$ form a covering for $P$ ($P \subseteq \bigcup_i X_i$)

• For every element $X_i$ of $U$ we cluster the points of $X_i$ with a suitable metric i.e for every $X_i$ we have the set of clusters $X_{ij}$;

• Every cluster $X_{ij}$ would be represented as a node in graph and if $X_{ij} \cap X_{rs} \neq \emptyset$ then we draw an edge between the nodes $X_{ij}$ and $X_{rs}$.

The intuitive idea behind Mapper is illustrated in figure 7 and can be explained as follows: suppose we have a point cloud data representing a shape, for example a "knot". First we project the whole data on a coordinate system with less dimensionality in order to reduce complexity via dimensionality reduction (here we project the data on the knot to x-axis). Now we partition the parameter space (x-axis) into several bins with an overlapping percentage. Next, put data into overlapping bins. Afterwards, we use clustering algorithms in order to classify the points of each bin into several clusters. Once the previous stage is done, we can create our interactive graph.

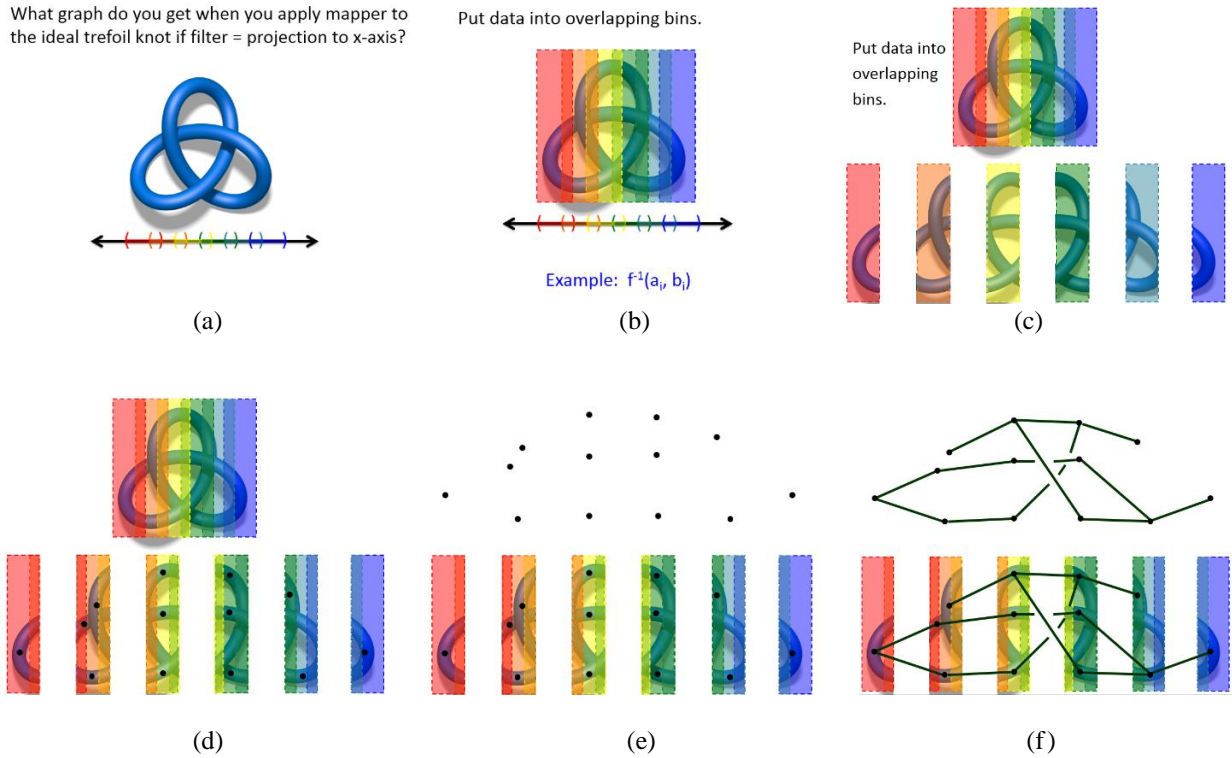

**Fig. 7:** Mapper algorithm on knot shape data cloud: **a)** First we project the whole data cloud to embedded space (here x-axis). **b)** Then we partition the embedded space into overlapping bins (here showed as colored intervals). **c)** Then we put data into overlapping bins. **d)** Next we use any clustering algorithm to cluster the points in the cloud data. **e&f)** each cluster of points in every bin would represented as a node of the graph and we draw and edge between two nodes if they share a common data point.

## 4.1 **TDA-Based Mapper Analysis online algorithm and our dataset**

Implementation of mapper algorithm is already avaliable in python packages like "Mapper" and " Kepler Mapper". In this article we used "Kepler Mapper" along side other python packages. In this paper, we used the textual data (poems) of two iranian poets Hafez and Ferdowsi. The data set gathered from "Shahnameh" (An epic book from Ferdowsi) and "Ghazaliat-e-Hafez", that includes about different 9000 hemistich (ranging from epic wars to love and romance) from both books. In our model as a first step we used " truncatedSVD" and "t-SNE" as filter functions after applying "TF-IDF" on our data. Next we choose resolution and overlap, there are many ways to choose resolution and overlapping percentage. The more resolution we have, the better data will be partitioned and classified and higher the overlapping percentage is, the more compact our resulting graph would be. For clustering we used "Agglomerative Clustering" avaliable in "sklearn" package with "cosine" similarity and complete linkage.

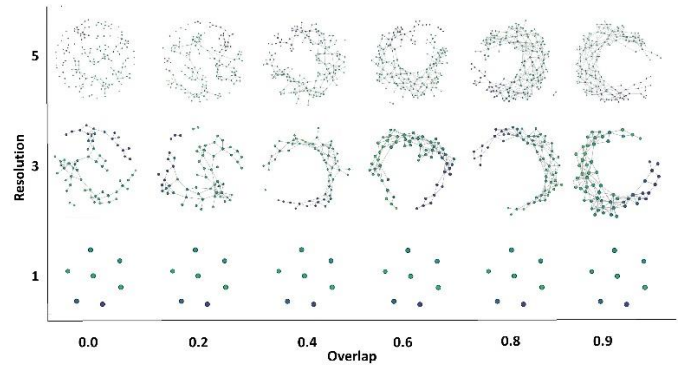

**Fig. 8**: Comparison of different resolutions and overlapping precentage: As is evident from the above table, the more resolution we have the better our resulting graph will be classified and the more the overlapping percentage is, the more compact the resulting graph would become.

## 4.2 **Results**

we examined two accuracy tests on our shape graph. First, we partition the whole graph into 3 clusters ("Hafez","Ferdowsi","Both"), Which in "Hafez" cluster we have the nodes which include the high percent of Hafezian poems, similarly in "Ferdowsi" cluster we have the nodes which include the high percent of poems of Ferdowsi and in the "Both" cluster we have about the same amount of both poems. We examine what percent of poems in "Hafez" cluster really belong to Hafezian poems and the same method for other clusters. To do this we simply divide the number of Hafezian poems in each node in the "Hafez" cluster by the number of all poems in each node in the same cluster, and we do the same test to other clusters as well. For example for "Hafez" cluster we have the following calculations:

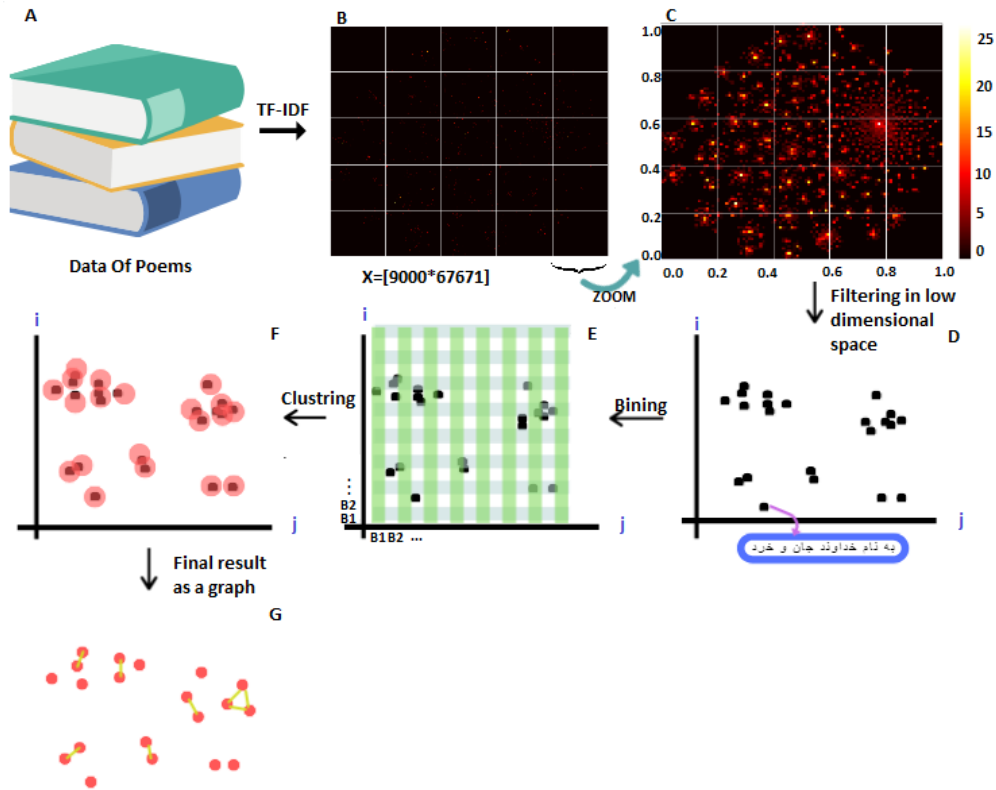

**Fig. 9**: TDA Mapper in text classification. A) Gathering and pre-processing the data of poems. B) Using TF-IDF algorithm to turn pre-processed data into a matrix which columns represent a single word in the entire corpus and each row represent a hemistich. The matrix is of the form [9000* 67671] (# hemistich *# words in corpus). C) Zoomed data [100*100]. D) Filteration step, Using truncated SVD and T-SNE

algorithms as filter functions. E) Binning: the whole 2-dimensional space divided in to smaller bins. F)
Clustering: clustering algorithms applied to each bin in order tocreate the nodes in the final graph. G)
Output of the algorithm as a graph, every node in the graph is a cluster of the previous step and nodes
share an edge if they have a data point in common.

**Percentage of accuracy = (Number of poems of Hafez in each node of the cluster / Number of all poems in the
whole cluster)**

.
So if we have the accuracy percentage of a for a cluster it means that a percent of the poems in that cluster has been
labelled correctly. second, we tried to devide some parts of the graph into several clusters based on their semantic
subjects and we created 5 clusters. To visually analyze each cluster werepresented the text with in each cluster as a
word cloud, which is easy to understand in just one sight.

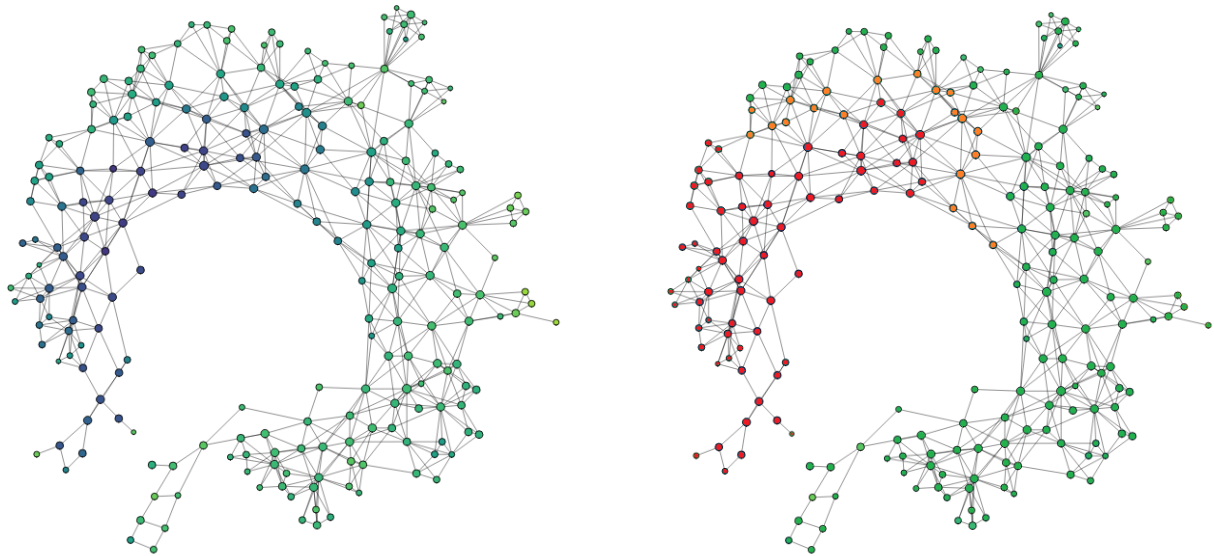

**(a) Original Graph**       **(b) Clustered for the first test**

**Fig. 10:** (a) shows the original output of the Mapper algorithm, while (b) shows the same graph which we
partitioned it into three clusters for the first test(Red for "Hafez" cluster, Orange for "Both" cluster, and
Green for "Ferdowsi" cluster). After examining the first test on each cluster, we got the following results:
for "Hafez" cluster percentage of accuracy was 80 percent, for"Ferdowsi" cluster percentage of accuracy
was about 94 percent ,and for "Both" cluster percentage of accuracy was 40 percent for Hafez poems and
60 percent for Ferdowsi poems. So for the "Hafez" cluster we can say that 80 percent of the poems in the
cluster has the rigth label and so on for other clusters.

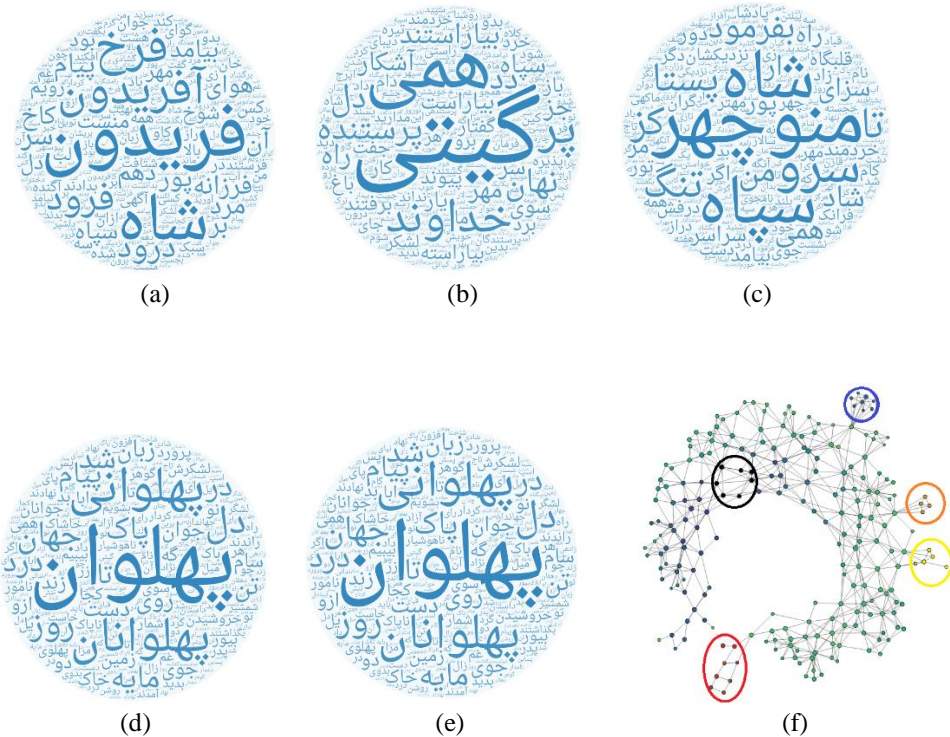

**Fig. 11:** semantic clustering of the graph, in image (f) we have examined the second test on our graph shape, we choose some clusters and then examined if they are semantically related to each other, we choosse 5 clusters ("Red","Blue","Black","Yellow",and "orange") as apparent in (f) and then draw their word cloud. (a) is word cloud of "Black" and it is clear that it belongs to Hafezian poems because of the frequency of some words and if we want to analyze it semantically it would be the cluster of romantic poems. (b) is word cloud for "Red" and because of the frequency of words in its word cloud we can say it semantically related to worshiping the god. (c) is word cloud for "Yellow". (d) is word cloud for "Orange".(e) is word cloud for "Blue"

## References

1. Bhattacharya, S., Ghrist, R., and Kumar, V. Persistent Homology for Path Planning in Uncertain Environments. IEEE TRANSACTIONS ON ROBOTICS, VOL. 31, NO. 3, JUNE 2015.

2. Bubenik, P., Dotko, P. A persistence landscapes toolbox for topological statistics. J. Symbol. Comput. 78, 91114 (2016).

3. Carlsson, G. Topology and data. Bull. Amer. Math. Soc, 46, 255308 (2009).

4. Chaski, C. Who's at the Keyword? Authorship Attribution in Digital Evidence Investigations. International Journal of Digital Evidence, Volume 4, Issue 1(2005).

5. Diederich, J., Kindermann, J., Leopold, E., Paas, G. Authorship Attribution with Support Vector Machines, Applied Intelligence, 19(1):109-123, 2003.

6. Edelsbrunner, H. Persistent Homology In Image processing. Lecture Notes in Computer Science book series.(LNCS, volume 7877), 2013.

7. Emrani, S., Saponas, T., Morris, D., and Krim, H. A. Novel Framework for Pulse Pressure Wave Analysis Using Persistent Homology. IEEE SIGNAL PROCESSING LETTERS, VOL. 22, NO. 11, NOVEMBER 2015.

8. Gholizadeh, S., Seyeditabari, A., and Zadrozny, W. Topological Signature of 19th Century Novelists: Persistent Homology in Text Mining, big data and cognitive computing. 2, 3, 2018, doi:10.3390/bdcc2040033.

9. Günther, D., Reininghaus, J., Hotz, I., and Wagner, H. Memory-Efficient Computation of Persistent Homology for 3D Images using Discrete Morse Theory. 24th SIBGRAPI Conference on Graphics, Patterns and Images, 2011.

10. Holmes, D. Authorship Attribution. Computers and the Humanities, 28:87-106. Kluwer Academic Publishers, 1995.

11. Jessica, L., Nielson, Jesse Paquette, Aiwen, W. Liu, Cristian F. Guandique, C., Amy. Tovar et al. Topological data analysis for discovery in preclinical spinal cord injury and traumatic brain injury. Nature Communications, Oct 14, 2015.

12. Lee, H., Chung, M. K., Kang, H., Kim B., Lee, D. DISCRIMINATIVE PERSISTENT HOMOLOGY OF BRAIN NETWORKS. 2011 IEEE International Symposium on Biomedical Imaging: From Nano to Macro.

13. Lee, H., Kang, H., Chung, M.K., Kim, B., and Lee, D. Persistent Brain Network Homology From the Perspective of Dendrogram. IEEE TRANSACTIONS ON MEDICAL IMAGING, VOL. 31, NO. 12, DECEMBER 2012.

14. Lum, P. Y., Singh. G, Lehman, A., Ishkanov. T., Vejdemo-Johansson, M., Alagappan, M., Carlsson, J., Carlsson, G. Extracting insights from the shape of complex data using topology, Sci. Rep, 3, 1236 (2013).

15. Lum, P. Y., Lehmann, L., Singh, G., Ishkhanov, T., Vejdemo-Johansson, M., Carlsson , G., The topology of politics: voting connectivity in the US House of Representatives. KTH, School of Computer Science and Communication (CSC), Computer Vision and Active Perception, CVAP, 2012.

16. Malyutov, M.B. Authorship Attribution of Texts: a Review. Proceedings of the program "Information transfer" held in ZIF. University of Bielefeld, Germany, 17 pages, 2004.

17. Maaten, L., Hinton, G. Visualizing data using t-SNE. J. Mach. Learn. Res. 9, 25792605 (2008).

18. Mihaela, E. Sardiu., Joshua, M. Gilmore., Groppe, B., Florens, L., Michael, P. Washburn. Identification of Topological Network Modules in Perturbed Protein Interaction Networks. Scientific Reports, 2017.

19. Nicolau, M., Levine, A. J., Carlsson, G. Topology based data analysis identifies a subgroup of breast cancers with a unique mutational profile and excellent survival. Proc Natl Acad Sci U S A, 108, 726570 (2011).

20. NILSSON, D., EKGREN, A., Topology and Word Spaces, Bachelors Thesis at CSC.

21. Rizvi, A., Camara, P., Kandror, E., Roberts, T., Schieren, I. et al. Single-cell topological RNA-seq analysis reveals insights into cellular differentiation and development, Nature Biotechnology, 2017.

22. Robles, A. , Hajij, M., Paul Rosen. The shape of an image: A study of mapper on images. arXiv:1710.09008v2 ,[cs.CV] 7 Dec 2017.

23. Romano, D., Nicolau, M., Quintin, E. M., Mazaika, P. K., Lightbody, A. A., Hazlett, H. C., Piven, J., Carlsson, G., Reiss, A. L. Topological methods reveal high and low functioning neuro-phenotypes within fragile X syndrome. Hum. Brain Mapp. 35, 49044915 (2014).

24. Saggar, M., Sporns, O., Gonzalez-Castillo, J., Bandettini, P., Carlsson, G. et al. Towards a new approach to reveal dynamical organization of the brain using topological data analysis, Nature Communications, Apr 11, 2018.

25. Singh, G., Mmoli, F., Carlsson, G. Topological methods for the analysis of high dimensional data sets and 3d object recognition. In Eurographics Symposium on Point-Based Graphics (eds Botsch, M., Pajarola, R.) (The Eurographics Association, 2007).

26. Stamatatos, E., Fakotakis, N., Kokkinakis, G. Computer-Based Authorship Attribution Without Lexical Measures. Computers and the Humanities, 35: 193-214, Kluwer Academic Publishers, 2001.

27. Torres P., Hromic H., Heravi B. Topic Detection in Twitter Using Topology Data Analysis. ICWE 2015 Workshops, LNCS 9396, pp. 186197, 2015.

28. Zhang, J., Ziyu, X., and Stan, Z. L. Prime Discriminant Simplicial Complex. . IEEE TRANSACTIONS ON NEURAL NETWORKS AND LEARNING SYSTEMS, VOL. 24, NO. 1, JANUARY 2013.

29. Zhu, X. Persistent Homology: An Introduction and a New Text Representation for Natural Language Processing. Twenty-Third International Joint Conference on Artificial Intelligence, 2013.